\documentclass[journal]{IEEEtran}
\usepackage{amsmath,amsfonts}
\usepackage{array}
\usepackage[caption=false,font=normalsize,labelfont=sf,textfont=sf]{subfig}
\usepackage{textcomp}
\usepackage{stfloats}
\usepackage{url}
\usepackage{verbatim}
\usepackage{wrapfig}
\usepackage{graphicx}
\usepackage{cite}
\usepackage[ruled]{algorithm2e}
\usepackage{todonotes}
\DeclareMathOperator*{\freak}{\|}

\usepackage{booktabs}
\usepackage{multirow}
\usepackage{amsmath}
\usepackage{mathrsfs}
\usepackage{amssymb}
\usepackage{graphicx} % Include figure files

\usepackage[linkcolor=blue,
            colorlinks,
            anchorcolor=blue,
            citecolor=green,
            ]{hyperref}
 
\usepackage[ruled]{algorithm2e}

\begin{document}

\title{Multimodal Graph Learning for Deepfake Detection}

\author{Zhiyuan Yan, Peng Sun, Yubo Lang, Shuo Du, Shanzhuo Zhang, Wei Wang, ~\IEEEmembership{Member,~IEEE}, Lei Liu
        % <-this % stops a space
\thanks{This work was supported by the Technical Research Program of the Ministry of Public Security [2020JSYJC25]; Open Project of Key Laboratory of Forensic Science of Ministry of Justice [KF202014].}
\thanks{Peng Sun (6094079@qq.com) and Yubo Lang (lang$\_$yubo@163.com) are with the School of Information Technology and Intelligence at the Criminal Investigation Police University of China in Shenyang, 110854, China. Zhiyuan Yan (yanzhiyuan1114@gmail.com), Shuo Du (shuodu052@gmail.com), and Lei Liu (380087534@qq.com) were previously at the same university. Zhiyuan Yan is now affiliated with the School of Data Science at the Chinese University of Hong Kong in Shenzhen, 518172, China, while Shuo Du is currently with the School of Control Science and Engineering at Dalian University of Technology in Dalian, 116031, China. Lei Liu is now with the School of Information Science and Engineering, Northeastern University, Shenyang, 110819, China.}
\thanks{Shanzhuo Zhang is with the Natural Language Processing Department of Baidu, Shenzhen, 518062, China (email: shanzhuo.zhang@gmail.com).}
\thanks{Wei Wang is with the School of Computer Science and Technology, Harbin Institute of Technology, Shenzhen, 518055, China (email: wangwei2019@hit.edu.cn).}
\thanks{ZY.Y and P.S formulate the overarching research goals and design the methodology. ZY.Y,  YB.L, and S.D write the initial draft. YB.L verifies the overall reproducibility of the experiments and visualizes the methodological processes. ZY.Y and S.D implement the computer code and supporting algorithms, test existing code components, and visualize the methodological processes. SZ.Z is involved in both designing the methodology and critically reviewing the original draft. P.S, W.W, and L.L verify the overall replication of the results and critically review the original draft.}
\thanks{The Corresponding author is Peng Sun (email: 6094079@qq.com)}}
% <-this % stops a space

\maketitle

\begin{abstract}
Existing deepfake detectors face several challenges in achieving robustness and generalization. One of the primary reasons is their limited ability to extract relevant information from forgery videos, especially in the presence of various artifacts such as spatial, frequency, temporal, and landmark mismatches.
%
% This creates a gap in the field of deepfake detection, as current detectors struggle to effectively identify and utilize distinguishing features, resulting in limited effectiveness in identifying deepfake videos.
%
Current detectors rely on pixel-level features that are easily affected by unknown disturbances or facial landmarks that do not provide sufficient information. Furthermore, most detectors cannot utilize information from multiple domains for detection, leading to limited effectiveness in identifying deepfake videos.
To address these limitations, we propose a novel framework, namely Multimodal Graph Learning (MGL) that leverages information from multiple modalities using two GNNs and several multimodal fusion modules. At the frame level, we employ a bi-directional cross-modal transformer and an adaptive gating mechanism to combine the features from the spatial and frequency domains with the geometric-enhanced landmark features captured by a GNN. At the video level, we use a Graph Attention Network (GAT) to represent each frame in a video as a node in a graph and encode temporal information into the edges of the graph to extract temporal inconsistency between frames.
Our proposed method aims to effectively identify and utilize distinguishing features for deepfake detection. 
We evaluate the effectiveness of our method through extensive experiments on widely-used benchmarks and demonstrate that our method outperforms the state-of-the-art detectors in terms of generalization ability and robustness against unknown disturbances.
\end{abstract}


\section{Introduction}

Deepfake videos have become a growing concern due to their potential to deceive and manipulate viewers. Unfortunately, these videos are often used to create misleading content that violates personal privacy, spreads false information, and undermines public trust in digital media. 

\begin{figure}[t!]
  \centering
  \includegraphics[width=8.5cm]{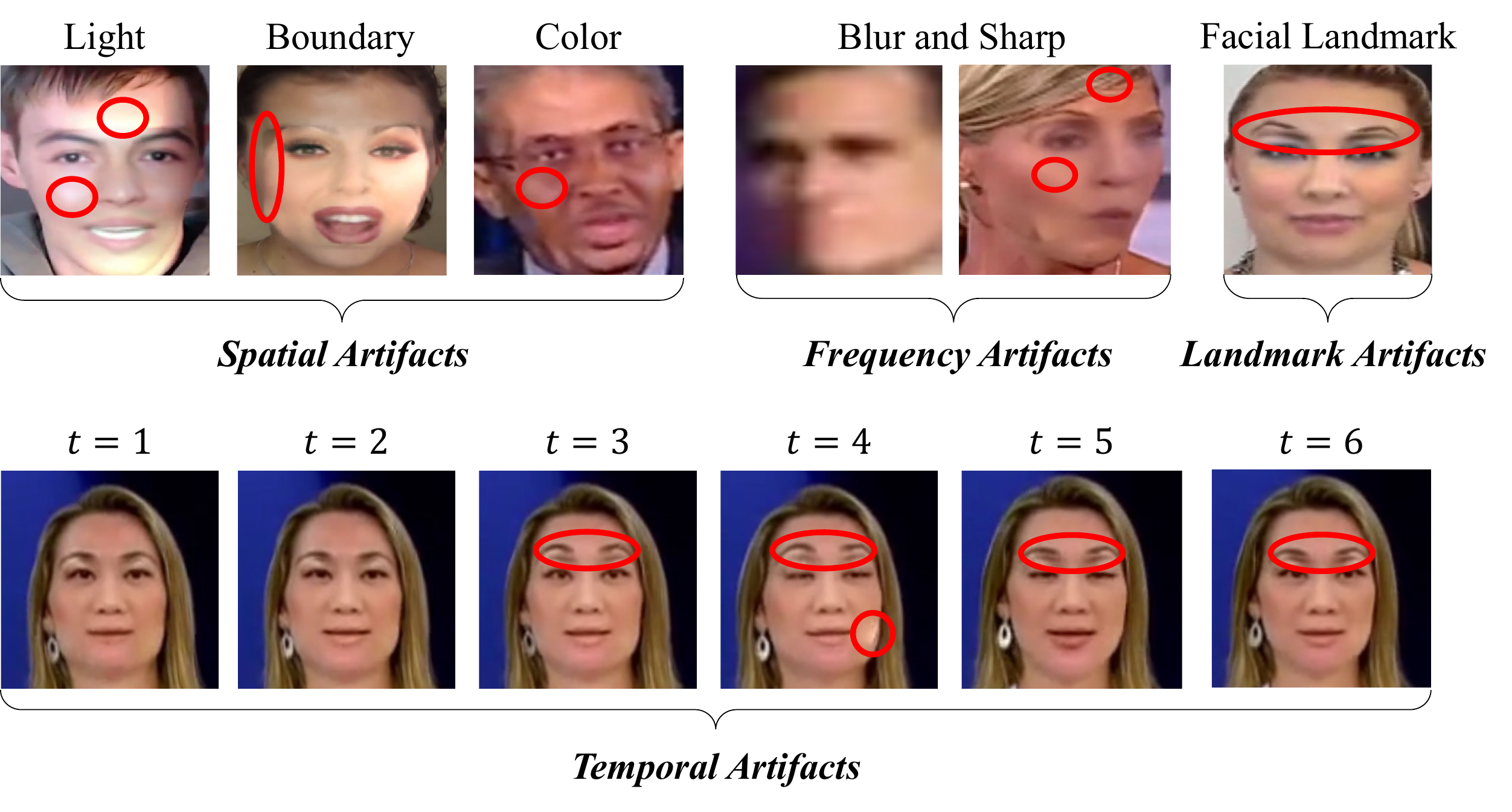}
  \caption{Deepfake synthesis techniques often generate artifacts that can be classified into spatial, frequency, landmark, and temporal artifacts. These artifacts can result in visual inconsistencies on the forgery faces.
  }\label{motivation}
\end{figure}

To mitigate these issues, reliable and effective deepfake detectors are needed. As such, deepfake detection has emerged as an important area of research. In recent years, there has been a growing interest in developing methods for detecting realistic videos~\cite{rossler2019faceforensics++,qi2020deeprhythm,li2020sharp,yu2022improving}. Early studies~\cite{mo2018fake,afchar2018mesonet,nguyen2019use} focus on designing optimal architecture networks that leverage spatial domain artifacts for detection. 
However, these spatial-based detectors often overlook artifacts in the frequency domain, such as recompression artifacts. As a result, some recent works~\cite{qian2020thinking,liu2021spatial} explore frequency domain features for detection.
Since deepfake videos are typically generated frame by frame, some methods~\cite{guera2018deepfake,sabir2019recurrent,gu2021spatiotemporal} detect deepfake by leveraging the temporal artifacts, \textit{i.e.,} inconsistency between frames.
However, these methods still experience generalization and robustness problems. The former refers to the detector's inability to identify testing data that have different distributions from the training data, while the latter refers to the detector's vulnerability to external perturbations, \textit{e.g.,} light, mask, and noise.

\begin{figure*}[htb]
      \centering
      \includegraphics[width=0.9\linewidth]{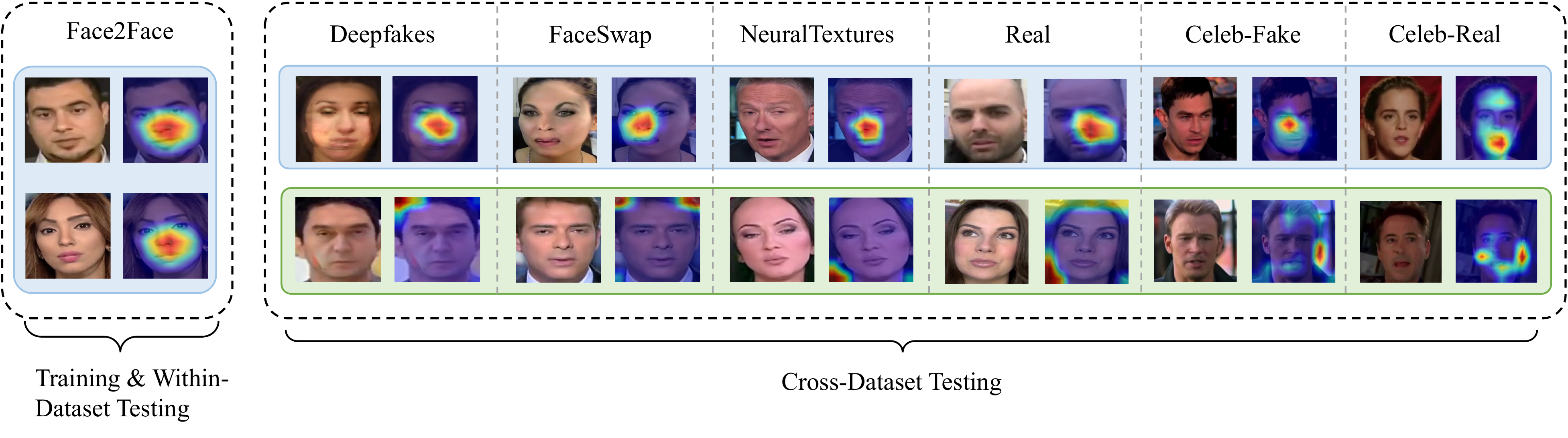} 
      \caption{
      Visualization of the baseline model (Xception) trained on Face2Face (F2F)~\cite{thies2016face2face} and tested on other unseen manipulation methods. We observe that there are two types of over-fitting. First, for images in the blue region, the model tends to over-fit the specific artifacts of F2F. Second, for the green region, the model over-fits to the forgery-irrelated textures (\textit{i.e.,} background and corner). These observations illustrate the limitations of the baseline in terms of generalization and robustness, as the model's responses during evaluation on unseen forgeries either resemble those in the real face or predict the same results as the F2F's images.
      }
      \label{fig:xcep_grad}
\end{figure*}

One reason for these problems is that current deepfake manipulation techniques can produce distinct artifacts, \textit{i.e.,} spatial, frequency, temporal artifacts, and landmark mismatch, as shown in Fig.~\ref{motivation}. However, most existing detectors are limited in their ability to identify these artifacts from the forgery video. Instead, as illustrated in Fig.~\ref{fig:xcep_grad}, the detectors can be overfitted to both the forgery-irrelevant features and method-specific features, thereby limiting their ability to learn generalizable and robust features required for effective deepfake detection. 

In addition, although these pixel-based features (\textit{i.e.,} spatial, frequency, and temporal features) can capture subtle facial information, \textit{e.g.,} color and texture, it is sensitively affected by external changes. Conversely, we observe that features of facial landmarks are more robust to such external perturbations since they record only the coordinates of different key points, which remain unchanged by external perturbations. Yet, facial landmarks are limited to capturing the subtle color and light variations in a synthetic image. As such, pixel-based and landmark-based features can complement each other in forgery detection. Thus, it is unknown whether combining features from different modalities, \textit{e.g.,} spatial, frequency, and temporal domain features with the facial landmarks, can improve the generalization and robustness of the model and how to effectively fuse these features from different modalities. This aspect is always overlooked by previous works.

To address these limitations, we propose a novel framework that leverages pixel-based features from spatial, temporal, and frequency domains, along with facial landmark features, to comprehensively represent face forgery video for detection. We also employ two GNNs for extracting the features at the frame and video levels, respectively. At the frame level, our approach differs from previous work that utilizes only low-dimension landmark coordinate information, which lacks sufficient details. Instead, we use a GNN to explicitly learn the geometric features of facial landmarks, resulting in a high-dimension and deeper representation. At the video level, we consider each frame of a video as a node in a graph and encode the similarity between frames as edge features. GNNs perform both local and global reasoning by aggregating information from neighbors and the overall graph structure, enabling the detector to capture both short-term and long-term inconsistencies in the forgery video. 

Overall, the major contributions in this paper are summarized as follows: 
\begin{itemize}
    \item We are the first work that extracts multimodal features from the spatial, frequency, and temporal domains, along with the facial landmark features for deepfake detection. In this manner, our framework aims to enhance both the generalization and robustness of the model for deepfake detection.
    \item We are the first work that utilizes GNNs from both the frame and video levels to detect video forgery. At the frame level, a GNN is utilized to learn a high-dimensional representation of facial landmarks, enhancing the robustness of the model. At the video level, a GNN is employed to learn both local and global consistencies in the forgery video by utilizing the edge features of the graph. Furthermore, to the best of our knowledge, no existing method utilizes GNNs to capture both the facial geometric and temporal forgery information for forgery detection, highlighting the need for further exploration in this area.
    \item We demonstrate that our method outperforms existing state-of-the-art and mainstream methods through extensive experiments and visualizations, providing further evidence of the effectiveness of our proposed framework.
\end{itemize}

\section{Related Work}
\label{relatedwork}

\subsection{Spatial-Based Deepfake Detector}
Due to the strong capability of feature extraction in the spatial domain, CNNs have become the mainstream detection method. Mo~\textit{et al.}~\cite{mo2018fake} leverage CNNs to identify forged images and achieve better results than the previous traditional methods. To further address the spatial inconsistencies created by the deepfake techniques in the forgery process, some works~\cite{he2019detection,li2018exposing} aim to locate these visual artifacts and identify fake images based on visual artifacts (such as color inconsistency, blending boundary, and blur artifacts). Afchar~\textit{et al.}~\cite{afchar2018mesonet} propose MesoNet combined with an Inception module to extract middle-level features for deepfake video detection. 
In recent years, deep learning techniques have been applied to enhance the performance of deepfake detection. Some recent works have explored new directions, such as the incorporation of vision transformers with incremental learning~\cite{khan2021video} and distillation~\cite{heo2021deepfake}. Other notable approaches have focused on specific representations, such as forgery region location~\cite{nguyen2019multi}, metric learning~\cite{cao2021metric}, and attentional networks~\cite{zhao2021multi,wangm2tr}. These works aim to extract more comprehensive and discriminative features for deepfake detection by leveraging different aspects of the input data.

However, forgery detection techniques that only rely on spatial artifacts often overlook the artifacts present in the frequency domain, such as re-compression artifacts. 
% Additionally, inconsistencies may exist at the high-frequency portion of the image, which can be challenging to identify using spatial domain methods. 

\subsection{Frequency-Based Deepfake Detectors}
Currently, an increasing number of studies explore frequency-based methods for forgery detection. For instance, Ricard~\textit{et al.}~\cite{RicardDurall2020WatchYU} demonstrate that convolution-based upsampling methods used in deepfake technologies can cause a mismatch in the spectral distribution between fake and real images and videos. To overcome this limitation, they propose a frequency-based scheme that outperforms most mainstream spatial-based methods. Furthermore, learning-based frequency domain methods are also well-studied. For example, Stuchi~\textit{et al.}~\cite{JosAugustoStuchi2017ImprovingIC} use filters to extract information in different ranges, followed by a fully connected layer to obtain the output. Qian~\textit{et al.}~\cite{qian2020thinking} design a set of learnable filters to adaptively mine frequency forgery clues using frequency-aware image decomposition.

However, detectors that rely on spatial and frequency artifacts may overlook the temporal inconsistency present in deepfake videos, which is essential to identify the real from the fake.

\subsection{Temporal-Based Deepfake Detector}
As deepfake videos are generated frame by frame, there are often differences between successive frames due to changes in lighting, noise, and motion. Detection methods that incorporate temporal information exploit this temporal incoherence to identify fake videos.
Currently, the mainstream approach to temporal detection is based on the CNN\_RNN structure. Sabir~\textit{et al.}~\cite{sabir2019recurrent} adopt a CNN\_RNN pipeline for deepfake detection, using the CNN to extract frame-level features and the RNN module to learn the temporal incoherence between frame sets. However, CNNs are more effective in modeling local patterns and may not perform well in capturing global patterns.
To address this limitation, Zheng~\textit{et al.}~\cite{zheng2021exploring} propose a temporal transformer to capture the long-term dependency between real and fake videos. However, transformers are designed to capture global dependencies and may not be as effective in capturing local information, which is crucial in detecting video forgery.

\subsection{Landmark-Based Deepfake Detectors}
Detectors that rely on pixel-level information, such as spatial, temporal, or frequency domains, are susceptible to perturbations caused by external factors and require substantial computational resources, making them difficult to deploy. 
In response to these challenges, researchers have focused on landmark-based detection methods that leverage facial landmarks to capture the movements of facial organs. For instance, Yang~\textit{et al.}~\cite{yang2019exposing} develop a detector based on head pose that distinguishes between real and fake videos by utilizing the spatial relationships of landmark information. Sun~\textit{et al.}~\cite{sun2021improving} apply deep neural networks to implicitly capture the relationship between different landmarks in both the spatial and temporal domains, demonstrating their effectiveness. 

However, these landmark-based detection methods are limited to using only landmark information and do not consider the fusion of information from different modalities. Furthermore, existing landmark-based methods do not explicitly capture the geometric facial landmarks to obtain a complex representation but only use the coordinate sequence, which lacks sufficient details.

\begin{figure*}[ht]
  \centering
  \includegraphics[ width=\linewidth]{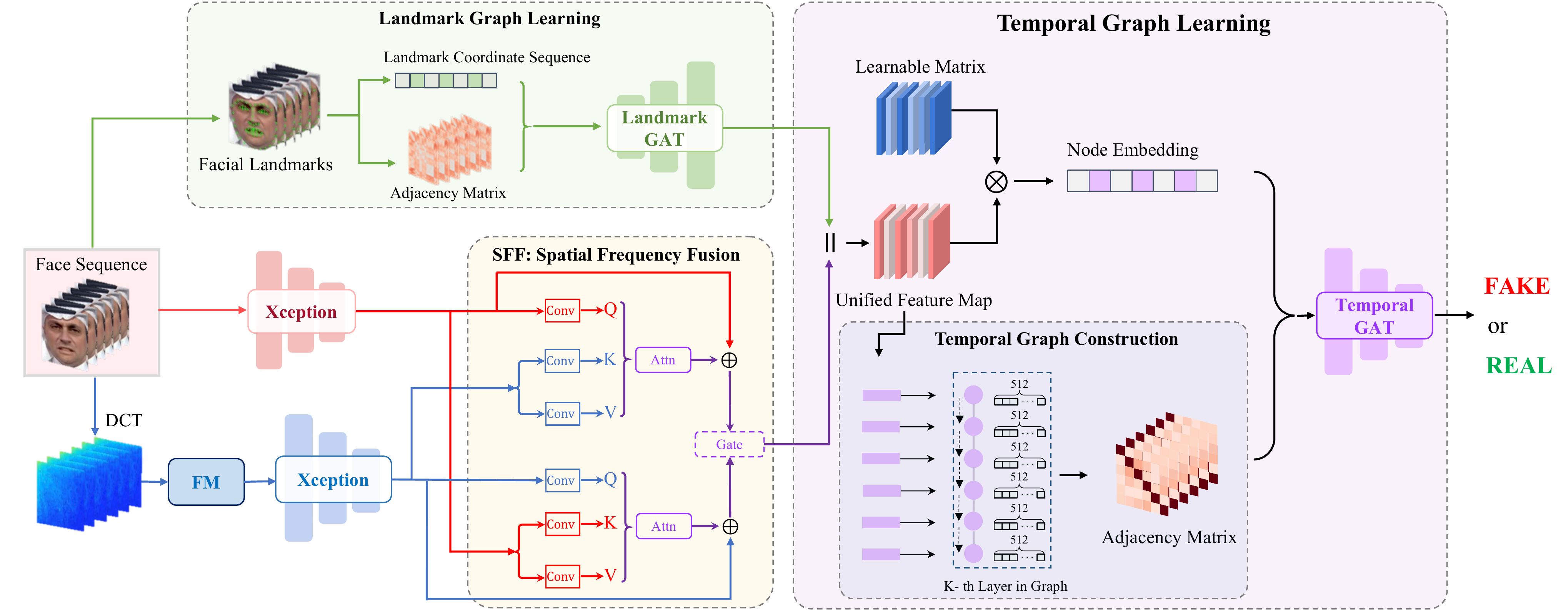}
  \caption{
  Overview of the pipeline of our proposed framework, where $\bigotimes$ donates matrix multiplication, $\bigoplus$ donates the element-wise addition, and $\parallel$ donates the concatenate operation. For convenience, we only demonstrate the pipeline using six frames in this figure.
  }\label{framework}
\end{figure*}

\section{Proposed Method}
\label{proposedmethod}

\subsection{Frame Level Features Extraction}
\label{FrameLevel}

\begin{figure}[t!]
  \centering
  \includegraphics[width=8.5cm]{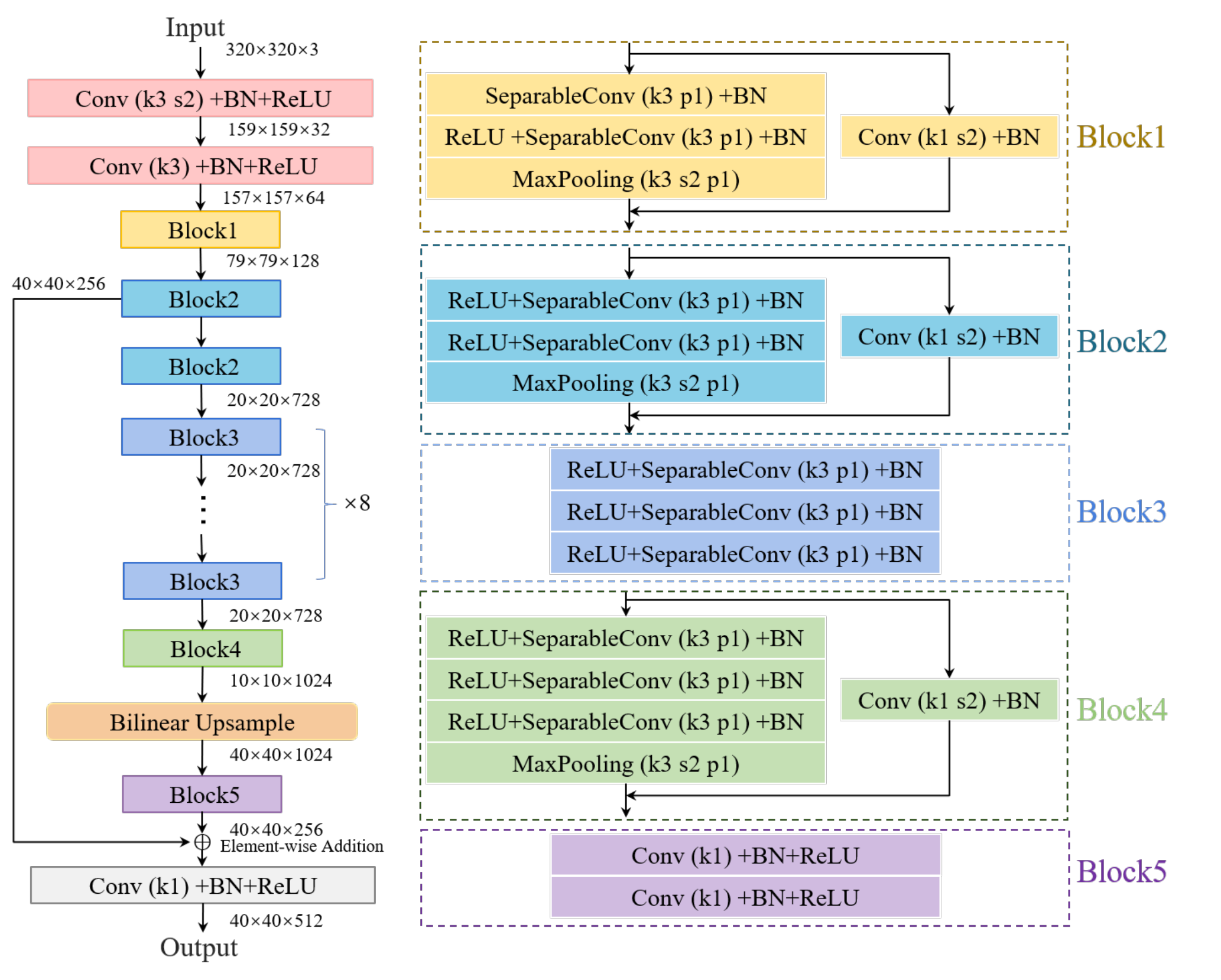}
  \caption
  {
  Overview of our improved Xception model architecture.
  }\label{improvedxcep}
\end{figure}

In this paper, a modified Xception architecture is employed for the extraction of spatial domain features, utilizing parameters pretrained in ImageNet. 
Given a pre-processed fragment $x$ $\in$ $\mathbb{R}^{C\times H\times W }$, where $C, H$ and $W$ are 3, 320, and 320, respectively, the output feature map at the final block has the shape $\mathbb{R}^{2048\times 7\times 7 }$ in the original Xception setting. 
It has been observed in previous studies that low-level features play a crucial role in deepfake detection. To address this, we redesign the architecture of Xception and develop an improved version with a multi-scale fusion module, as illustrated in Fig.~\ref{improvedxcep}.
Our improved Xception combines the feature maps from block2 and block5, which are subsequently passed through a convolution layer with a kernel size of 1 to obtain the multi-scale features $X_{s}$. It is important to note that $X_{s}$ $\in$ $\mathbb{R}^{512\times 40\times 40 }$, with a channel size of 512, which is four times smaller than that of the original Xception (2048).

\paragraph{Frequency Features Extraction}
\label{frequencydomain}
To transform the input $x$ from the spatial domain to the frequency domain, the Discrete Cosine Transform (DCT) is adopted in this paper. 

To remove the redundant information in the frequency domain and reduce noise interference, we use a binary mask $M$ $\in$ $\mathbb{R}^{320 \times 320}$, which is a symmetric square matrix:

\begin{equation}
%$\{ M | M_{ij}=1 \quad if \quad  up < i+j <low ；\quad else \quad M_{ij}=0 \}$, 
    M:M_{ij}=\begin{cases}1, \quad \textit{if} \quad \textit{low} < i+j < \textit{up} \\
                            0,\quad \textit{otherwise} \end{cases},
\label{eq1}
\end{equation}

\noindent
where $low$ is the lower cutoff frequency, and $up$ is the upper cutoff frequency.

\begin{figure}[t!]
  \centering
  \includegraphics[width=8.5cm]{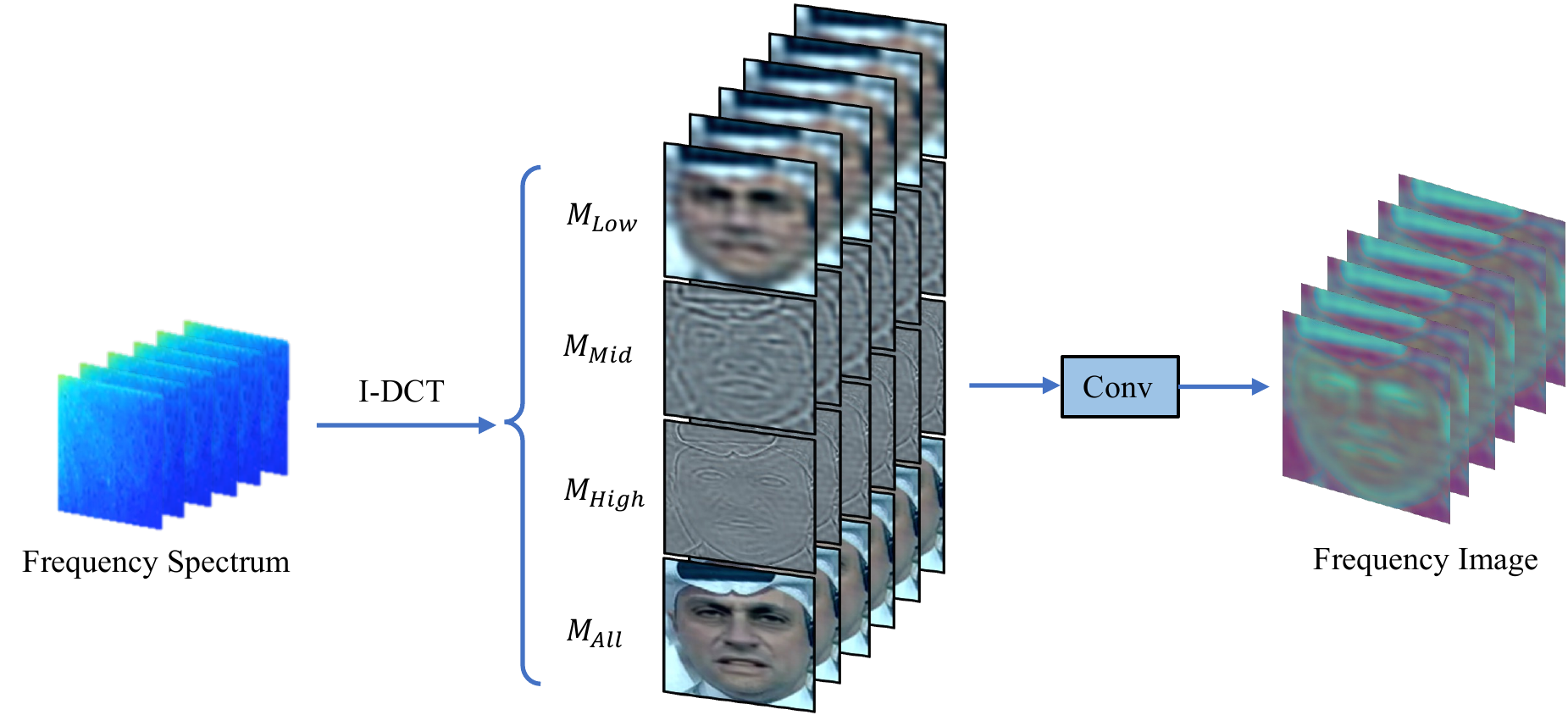}
  \caption
  {
  Overview of the Frequency Module (FM) utilized in our proposed method, which involves applying an inverse Discrete Cosine Transform (I-DCT) to the input image. We then apply four different filters ($M_{Low}$, $M_{Mid}$, $M_{High}$, $M_{All}$) to obtain separate frequency components. The outputs from these filters are then combined using a convolutional layer to synthesize the frequency image with 3 channels.
  }
  \label{frequencymodule}
\end{figure}

Previous efforts~\cite{qian2020thinking,wang2021m2tr} show that the spectrum of a face image is most efficiently extracted with filters in three different bands: low, medium, and high. Note that we also use a residual structure to add the original frequency information and add a corresponding learnable mask $M_{s}$ $\in$ $\mathbb{R}^{320 \times 320}$ to each band to extract forgery information in an adaptive manner:
\begin{equation}
\begin{aligned}
        y_{f} = \mathscr{D}\left( x\right) & \odot \begin{pmatrix}
        M_{low}+\sigma \left( M_{s}^{l}\right) \\
        M_{mid}+\sigma \left( M_{s}^{m}\right) \\
        M_{hig}+\sigma \left( M_{s}^{h}\right) \\
        M_{all}+\sigma \left( M_{s}^{a}\right)
\end{pmatrix}, \\
\end{aligned}
\label{eq2}
\end{equation}

\noindent
where $y_{f}$ denotes features after frequency domain transformation, $\mathscr{D}$ represents DCT, $\odot$ denotes the element wise dot-product, $M_{all}$ is an all-pass mask $\{M_{all} \; | \;  M_{ij}=1\}$, and $\sigma\left( \cdot  \right)$ denotes sigmoid function. 
To align the shape of the spatial features and obtain the distinctive frequency features, the FM module is employed (see Fig.~\ref{frequencymodule}). The output of the FM module is donated as $X_f$. The spatial and frequency Xception share the same structure, but they do not share parameters.
Also, since the shape of $y_{f}$ is $\mathbb{R}^{12 \times 320 \times 320}$, whose channel size (12) is four times that of $x$ (3), we also apply a pointwise convolution before the frequency Xception to adjust the channel size from 12 to 3.

\paragraph{Spatial-Frequency-Fusion (SFF) Module}
\label{sff}

The SFF module takes in two input feature maps, spatial and frequency, and passes them through two separate convolutional layers to extract features, respectively. The extracted features are then fed into a Cross-Modal Transformer (CMT) that applies the self-attention mechanism to the spatial and frequency features to learn cross-modal dependencies. Finally, an adaptive fusion mechanism is used to combine the features from the two modalities.

The CMT module consists of a multi-head self-attention layer and a feedforward neural network with residual connections and layer normalization. The output of the multi-head self-attention layer is given by:

\begin{equation}
\begin{aligned}
    Z_{s} &= \textit{MHA}(X_{s}, X_{f}, X_{f}) + X_{s}, \\
    Z_{f} &= \textit{MHA}(X_{f}, X_{s}, X_{s}) + X_{f},
\end{aligned}
\label{eq3}
\end{equation}
where $Z_{s}$ and $Z_{f}$ denote the output of the CMT. Also, the multi-head attention mechanism is denoted by $\textit{MHA}$. The bi-directional self-attention mechanism is used in the multi-head attention layer denoted by $\textit{MHA}$. It is applied twice in the cross-modal transformer. In the first application, the input $X_{s}$ serves as the query, while $X_{f}$ serves as the key and value. In the second application, $X_{f}$ is the query, while $X_{s}$ is the key and value.

Finally, gating coefficients $G_{s}$ and $G_{f}$ are used to fuse the spatial and frequency features using an adaptive mechanism. The unified features are obtained by element-wise multiplication of the gating coefficients and the respective feature maps:

\begin{equation}
    X_{sff} = G_{s} \odot Z_{s} + G_{f} \odot Z_{f},
\label{eq4}
\end{equation}
where $\odot$ denotes element-wise multiplication.

\paragraph{Landmark Graph Learning (LGL) Module}
\label{landmark}
To enhance the robustness of the model, we further incorporate information on facial landmarks.  
% This is because the geometric relationship between facial organs is considered to be stable and less susceptible to external factors such as lighting and color. However, relying solely on the facial landmark sequence is straightforward and naive.
Specifically, we construct a graph to encapsulate the geometric relationship between facial organs, which is then transformed into a high-dimensional representation. This representation is fused with unified features obtained from the spatial and frequency domains at the frame level.

Formally, let $L \in \mathbb{R}^{B\times 68\times 2}$ denote the input landmarks for a batch of size $B$. Let $A \in \mathbb{R}^{B\times 68\times 68}$ denote the adjacency matrix, which is computed by computing the Euclidean distance between each pair of landmarks. The landmark graph neural network applies two graph attention layers (GATs) to the landmark nodes to obtain an updated graph-level representation. Each GAT consists of a linear projection followed by an attention mechanism that computes attention coefficients based on the pairwise relationships between the landmark nodes. Specifically, let $H^{(i)} \in \mathbb{R}^{B\times 68\times d_i}$ denote the hidden feature representation at layer $i$ with $d_i$ as the dimension of the hidden features. The GAT updates $H^{(i)}$ as follows:

\begin{equation}
H^{(i)} = ReLu \left( \sum_{j=1}^{68} \sum_{k=1}^{68} \alpha_{j,k}^{(i)} \cdot \textit{Linear}(H_j^{(i-1)}) \right),
\label{eq5}
\end{equation}
where $ReLu$ denotes the ReLU activation function, $\alpha_{j,k}^{(i)}$ denotes the attention coefficient for landmark $j$ and landmark $k$ at layer $i$, which is computed based on the hidden feature representations $H^{(i-1)}$. The adjacency matrix $A$ is used to weigh the pairwise relationships between the landmarks, such that the attention coefficients $\alpha_{j,k}^{(i)}$ are higher for pairs of landmarks that are more strongly connected in the graph. Finally, the graph-level representation is obtained by averaging the updated feature representations across all the landmarks. The final output of the landmark GAT can be obtained by:

\begin{equation}
X_{lmk} = \frac{1}{68} \sum_{j=1}^{68} H_j^{(N)},
\label{eq6}
\end{equation}
where $N$ is the number of layers in the landmark GAT.

\paragraph{Multimodal Feature Fusion}
\label{multimodal}
The multimodal fusion approach we propose involves learning the joint representations from multimodal inputs at the frame level. The landmark modality $X_{lmk}$ provides the shape, contour, and position of facial important points, whereas the spatial-frequency-domain modality $X_{sff}$ combines feature maps in the spatial and frequency domain.

To fuse the two modalities' features, we concatenate them and use a learnable weight matrix $M_{l}$ to extract the fusion features from the combined modality.

\begin{equation}
    X_{u} = \left( X_{sff} \parallel X_{lmk} \right) \odot M_{l},
\label{eq7}
\end{equation}
where $X_{u}$ $\in$ $\mathbb{R}^{512\times40\times40}$ donates the joint representation of the spatial, frequency, and landmark modalities.
It will be then passed through a max pooling layer to obtain the final combined representation of multimodal inputs at the frame level.

\subsection{Video Level Features Extraction}
\label{videolevel}

In this section, we present the proposed method for extracting temporal features from video frames. We treat video frames as nodes in a graph and model their temporal relationships using a graph-based approach. Specifically, we employ a Graph Neural Network (GNN) consisting of multiple Graph Layers to learn the temporal dependencies between frames. This is achieved by constructing a graph with the video frames as nodes and their pairwise similarities as edge features, followed by applying a Multi-Head Graph Attention mechanism to capture the temporal dependencies.
Note that the features for each node are initialized from the final combined representation at the frame level.

\paragraph{Overall GNN Architecture}
The GNN architecture consists of multiple Graph Layers, with each layer, employing a Multi-Head Graph Attention mechanism. The number of layers and heads can be adjusted to model the temporal dependencies between video frames. After passing through the GNN layers, we apply an Attention Module to obtain the final video-level features.

\paragraph{Edge Feature Formulation}
We represent the relationship between video frames by encoding their pairwise similarity as edge features. Formally, given an input feature $F \in \mathbb{R}^{B \times T \times D}$, where $B$ is the batch size, $T$ is the number of frames, and $D$ is the feature dimension, we first normalize the feature vectors to have unit length. Then, we compute the cosine similarity between all pairs of normalized feature vectors, forming an edge feature $E \in \mathbb{R}^{B \times N \times N}$, where $N = T$ is the number of nodes in the graph. The entry $E_{ij}$ in the edge feature corresponds to the cosine similarity between frames $i$ and $j$, capturing the pairwise similarity between frames and providing the basis for modeling the graph structure.

\paragraph{Graph Layer}
\label{graphlayer}
The Graph Layer consists of a Multi-Head Graph Attention mechanism, Layer Normalization, and a feed-forward network. It takes as input the node features $X$, the adjacency matrix $A$, and the edge features $E$, and returns a feature $H \in \mathbb{R}^{B \times N \times D}$ encoding the transformed node features.

The Multi-Head Graph Attention mechanism is defined as:

\begin{equation}
H_{i}=\left( \freak\limits_{k=1}^{m} \textit{head}_k(X, A, E)\right) W^O,
\label{eq8}
\end{equation}
where $\textit{head}_k$ denotes the graph attention head corresponding to the $k$-th attention mechanism, and $m$ is the number of attention mechanisms. Each attention head computes attention coefficients using the following formula:

\begin{equation}
\alpha_{i,j}^k = \textit{SoftMax}\Big(\frac{1}{\sqrt{D}}\Big(\textit{W}^Q_k X_i\Big)^\top \Big(\frac{1}{\sqrt{D}}\textit{W}^K_k X_j + \textit{E}_{i,j}\textit{W}^V_k\Big)\Big),
\label{eq9}
\end{equation}
where $\textit{W}^Q_k$, $\textit{W}^K_k$, and $\textit{W}^V_k$ are learnable weight matrices for the $k$-th attention mechanism, and $X_i$ denotes the feature vector of node $i$. The edge features $E_{i,j}$ are used as additional inputs to the attention computation. The attention coefficients are used to compute a weighted sum of the node features:

\begin{equation}
h_i^k = \sum_{j=1}^N \alpha_{i,j}^k \Big(\frac{1}{\sqrt{D}}\textit{W}^V_k X_j\Big).
\label{eq10}
\end{equation}

The output of each attention head is concatenated and passed through a linear layer with learnable weights $W^O$ to produce the output $H$. The resulting node features in the output $H$ effectively capture the temporal dependencies between video frames.

After the Multi-Head Graph Attention mechanism, the Graph Layer applies Layer Normalization and a feed-forward network with ReLU activation to further refine the node features.

\paragraph{Attention Module}
The Attention Module computes a weighted sum of the node features using attention weights, effectively aggregating the node features into a single vector representing the entire video, donated as $X_{gat}$. Our proposed GNN architecture aims to capture and preserve both local and global temporal dependencies in the video frames.

\subsection{Optimization Objective}
\label{loss}
In the last part, we apply an MLP layer to extract the deeper semantic information for the final features, which includes several $3 \times 3$ convolutional layers and normalization layers. Then, the final predicted label can be calculated by the softmax function:

\begin{equation}
    \hat{Y} = \textit{SoftMax}\left(\textit{MLP}\left( X_{gat} \right)\right).
\label{eq11}
\end{equation}

We adopt the cross-entropy loss function $\mathscr{L}_{c}$ to optimize the model. Finally, the parameters of the network are updated via back-propagation. The overall procedure of our proposed MGL can be seen in Alg.~\ref{alg}.

\begin{algorithm}
\label{alg}
\small
\caption{Training Procedure for MGL}
\LinesNumbered
\KwIn{Facial frames $\left\{ x_{1}, x_{2}, ..., x_{n} \right\}$}
\KwOut{Model parameters $\alpha = \left\{ w_{1}, w_{2}, b_{1}, b_{2}, ... \right\}$}

Randomly initialize $\alpha$\;
Randomly initialize the parameters of DCT filters;

\For{each batch in the training set}{
Extract spatial feature $X_{s}$ using the multi-scale Xception\;
\For{filter in filter set}{
Compute and fuse features in different frequency bands to obtain $y_{f}$ using Eqs.~\ref{eq1}, \ref{eq2};
}

Acquire the unified feature $X_{sff}$ for spatial and frequency features using Eqs.~\ref{eq3}, \ref{eq4} in the SFF module\;

Obtain high-dimensional representation of landmark geometric feature $X_{lmk}$ using Eqs.~\ref{eq5}, \ref{eq6}\;

Fuse and acquire the final representation at frame level $X_{u}$ using a learnable matrix and Eq.~\ref{eq7}\;

Compute attention coefficients of temporal GAT $\alpha_{i,j}$ between two nodes $i,j$ using Eq.~\ref{eq9}\;

Calculate edge feature $E_{i,j}$ between two nodes $i,j$ using cosine similarity\;

Update and obtain representation of temporal GAT $H_{i}$ using Eqs.~\ref{eq8}, \ref{eq10}\;

Acquire representation of the whole graph structure $X_{gat}$ using the attention mechanism\;

Predict $\hat{Y}$ using Eq.~\ref{eq11}\;

Compute loss $\mathscr{L}_{c}$ and update parameters $\alpha$ according to the gradient of $\mathscr{L}_{c}$\;
}
\Return $\alpha$;
\end{algorithm}

\begin{table*}[tb!]
\centering
\caption{Comparison of our proposed method with eight other competing methods on the FF++ dataset. We evaluate the performance of each method using the AUC, ACC, and EER metrics at different compression levels.}
\setlength{\tabcolsep}{1.0mm}{
\begin{tabular}{c|ccccc|ccccc|ccccc}
\toprule
\multirow{2}{*}{Metric} & \multicolumn{5}{c|}{AUC (Higher is better)} & \multicolumn{5}{c|}{ACC (\%) (Higher is better)} & \multicolumn{5}{c}{EER (Lower is better)} \\
\cmidrule(lr){2-6} \cmidrule(lr){7-11} \cmidrule(lr){12-16}
Dataset & DF & F2F & FS & NT & AVG & DF & F2F & FS & NT & AVG & DF & F2F & FS & NT & AVG \\
\midrule
\multicolumn{16}{c}{FF++ High Quality (c23)} \\
\midrule
HeadPose~\cite{yang2019exposing} & 0.678 & 0.568 & 0.533 & 0.501 & 0.551 & 61.77 & 56.81 & 53.26 & 50.10 & 55.49 & 38.910 & 43.790 & 46.900 & 49.910 & 44.880 \\
FDFClassifer~\cite{BinhMLe2022ExploringTA} & 0.481 & 0.492 & 0.496 & 0.529 & 0.499 & 51.28 & 49.22 & 49.58 & 52.80 & 50.72 & 51.270 & 50.530 & 50.270 & 48.080 & 50.040 \\
Xception~\cite{chollet2017xception} & 0.993 & 0.993 & 0.995 & 0.971 & 0.988 & 95.75 & 97.04 & 97.57 & 90.92 & 95.32 & 4.241 & 2.701 & 2.812 & 9.442 & 4.799 \\
MesoNet~\cite{afchar2018mesonet} & 0.836 & 0.601 & 0.619 & 0.674 & 0.632 & 74.21 & 56.33 & 58.10 & 59.63 & 62.07 & 24.498 & 43.661 & 41.217 & 37.143 & 36.630 \\
Meso-Incep~\cite{afchar2018mesonet} & 0.984 & 0.904 & 0.946 & 0.589 & 0.632 & 92.99 & 81.72 & 80.63 & 56.57 & 78.00 & 6.763 & 17.299 & 12.991 & 43.170 & 20.056 \\
CapsuleNet ~\cite{nguyen2019capsule} & $0.987$ & $0.984$ & $0.984$ & $0.940$ & $0.974$ & $95.28$ & $94.49$ & $94.22$ & $87.32$ & $92.83$ & $4.780$ & $5.710$ & $5.710$ & $12.50$ & $7.180$\\
CNN\_RNN~\cite{guera2018deepfake} & $0.987$ & $0.984$ & $0.973$ & $0.927$ & $0.968$ & $94.21$ & $94.86$ & $91.56$ & $84.86$ & $91.37$ & $5.893$ & $5.357$ & $8.460$ & $15.134$ & $8.711$ \\
F3Net~\cite{qian2020thinking} & $0.998$ & $0.990$ & $\mathbf{0.996}$ & $0.980$ & $0.991$ & $97.88$ & $97.94$ & $98.20$ & $93.49$ & $96.76$ & $2.076$ & $1.987$ & $2.165$ & $7.009$ & $3.309$ \\
\textbf{Ours} & $\mathbf{0.999}$ & $\mathbf{0.996}$  & $0.995$ & $\mathbf{0.987}$ & $\mathbf{0.994}$ & $\mathbf{98.13}$ & $\mathbf{98.13}$ & $\mathbf{98.93}$ & $\mathbf{95.27}$ & $\mathbf{97.62}$ & $\mathbf{1.786}$ & $\mathbf{1.786}$ & $\mathbf{1.488}$ & $\mathbf{4.643}$ & $\mathbf{2.426}$\\
\midrule
\multicolumn{16}{c}{FF++ Low Quality (c40)} \\
\midrule
HeadPose~\cite{yang2019exposing} & $0.482$ & $0.496$ & $0.474$ & $0.534$ & $0.497$ & $48.21$ & $49.63$ & $47.43$ & $53.43$ & $49.68$ & $51.150$ & $50.240$ & $51.62$ & $47.480$ & $50.120$\\
FDFClassifer~\cite{BinhMLe2022ExploringTA} & $0.599$ & $0.600$ & $0.549$ & $0.490$ & $0.550$ & $59.81$ & $55.94$ & $54.92$ & $49.01$ & $54.92$ & $40.490$ & $44.890$ & $45.20$ & $50.810$ & $45.350$\\
Xception~\cite{chollet2017xception} & $0.961$ & $0.901$ & $0.945$ & $0.773$ & $0.895$ & $89.41$ & $81.72$ & $86.81$ & $69.78$ & $81.92$ & $10.692$ & $18.661$ & $13.281$ & $30.312$ & $18.237$ \\
MesoNet~\cite{afchar2018mesonet} & $0.546$ & $0.658$ & $0.615$ & $0.557$ & $0.632$ & $48.50$ & $58.11$ & $57.33$ & $54.21$ & $54.54$ & $47.350$ & $39.560$ & $41.090$ & $46.280$ & $43.570$\\
Meso-Incep~\cite{afchar2018mesonet} & $0.957$ & $0.785$ & $0.747$ & $0.704$ & $0.798$ & $89.19$ & $70.70$ & $67.56$ & $64.49$ & $72.99$ & $10.879$ & $29.688$ & $32.344$ & $35.491$ & $27.101$\\
CapsuleNet~\cite{nguyen2019capsule} & $0.953$ & $0.879$ & $0.922$ & $0.810$ & $0.891$ & $88.51$ & $80.58$ & $85.26$ & $73.56$ & $81.98$ & $11.230$ & $19.480$ & $14.490$ & $25.670$ & $17.720$\\
CNN\_RNN~\cite{guera2018deepfake} & $0.946$ & $0.856$ & $0.946$ & $0.790$ & $0.885$ & $88.16$ & $76.98$ & $86.99$ & $70.16$ & $80.57$ & $11.875$ & $23.058$ & $13.058$ & $29.241$ & $19.308$ \\
F3Net~\cite{qian2020thinking} & $0.984$ & $0.952$ & $0.976$ & $0.854$ & $0.942$ & $93.04$ & $87.60$ & $93.13$ & $77.40$ & $87.79$ & $6.320$ & $12.001$ & $7.031$ & $22.366$ & $11.930$ \\
\textbf{Ours} & $\mathbf{0.996}$ & $\mathbf{0.970}$  & $\mathbf{0.978}$ & $\mathbf{0.920}$ & $\mathbf{0.966}$ & $\mathbf{96.60}$ & $\mathbf{92.77}$ & $\mathbf{93.57}$ & $\mathbf{83.04}$ & $\mathbf{91.50}$ & $\mathbf{3.390}$ & $\mathbf{8.155}$ & $\mathbf{6.786}$ & $\mathbf{17.143}$ & $\mathbf{8.869}$\\
\bottomrule
\end{tabular}}
\label{baseline}
\end{table*}

\section{EXPERIMENTS}
\label{experiment}

\subsection{Experiment Settings}
\label{setting}
\paragraph{Datasets}
\label{dataset}
During the research process of deepfake detection, several challenging datasets have been released. In this paper, we adopt two widely used datasets in deepfake detection in our experiments, \textit{i.e.,} FaceForensic++ (FF++)~\cite{rossler2019faceforensics++} dataset, DeepfakeDetection (DFD)~\cite{dfd}, and CelebDF~\cite{li2020celeb}. FF++ contains 1000 original videos and each video has three versions, namely the original version (raw), slightly-compressed version (c23), and heavily-compressed version (c40). Since videos in the real scenario have limited resolutions, we conduct all experiments in the settings of low (c40) and high (c23) compression. FF++ also contains four manipulated methods, including Deepfakes (DF), Face2Face (F2F), FaceSwap (FS), and NeuralTextures (NT). 

For the generalization ability evaluation, following~\cite{zhao2021multi,liu2021spatial}, we train our model on FF++ and evaluate it on CelebDF and DFD (unseen data). By default, the FF++ (c23) is adopted for evaluating the generalization ability. If there is any deviation from this default, it will be explicitly stated. CelebDF is a widely used testing dataset with high visual quality, which contains 5639 fake videos and 540 real videos. DFD~\cite{dfd} is a database released by Google, which contains 363 source videos from 28 actors and about 3,000 forged videos.

% \paragraph{Video Sampling Strategy}
% \label{sample}
% % Because of the expensive cost of sampling too many frames for training (\textit{i.e.,} 270 frames), 
% In our paper, we only select 32 frames of each video for training and testing. To obtain more diverse training data, we shuffle all the frames in the video and then sort the first 8 frames, thus ensuring that we sample 8 randomly spaced frames in sequence. By repeating the above procedure 4 times, we can use 32 frames in one video for training. In the rest of this paper, we use a function $\textit{R}\left( \cdot  \right)$ to donate the video sampling strategy.

% \paragraph{Data Augmentation}
% \label{preprocess}
% In this paper, the following augmentations are considered: (\textit{i}) BC: Brightness and Contrast changes (\textit{ii}) HSV: Hue, Saturation and Value changes (\textit{iii}) GB: Gaussian blur (\textit{iv}) JPEG: JPEG compression with a random quality factor between 50 and 99. The RGB images of the extracted face regions are resized to $3\times 320\times 320$ to obtain the face sequence $\left [x_{1},x_{2},...,x_{n} \right ]$.

\paragraph{Implementations}
\label{params}
In our model, we set the number of layers in the two GAT models to 5. For the GAT of the landmark, we set the number of input channel and output channels to 32 and 64, respectively. For the temporal GAT, we set the input channel, hidden channel, and number of heads to 512, 512, and 5, respectively.
For the optimization process, we use the AdamW optimizer \cite{loshchilov2018fixing}, with an initial learning rate of 0.00008. The batch size is fixed at 8 for both the training and testing procedures. We train the model for up to 50 epochs until convergence.
Furthermore, we apply commonly used data augmentations, such as image compression and horizontal flip.

\paragraph{Baseline Methods}
\label{baseline}
Our model is compared with several baseline methods that are trained with the settings used in the original papers. The first method, Headpose~\cite{yang2019exposing}, is a traditional machine-learning model that detects manipulated videos by estimating 3D head orientations and using the inconsistencies in head poses. It employs an SVM classifier for classification. The second baseline, FDFClassifier~\cite{BinhMLe2022ExploringTA}, is also a traditional machine learning method that uses the spectrum of an image as features and then trains the dataset with the SVM model and a hybrid Gaussian model to obtain the classification results. Xception~\cite{chollet2017xception} is another baseline, which is a CNN-based classifier widely used in deepfake detection works due to its high performance on relative benchmarks. MesoNet~\cite{afchar2018mesonet} is a CNN-based baseline method that uses fewer parameters than Xception and other CNNs with comparable performance. Meso-Incep~\cite{afchar2018mesonet} is the modification model that uses the InceptionNet to the original MesoNet. CapsuleNet~\cite{nguyen2019capsule} is another CNN-based baseline, which employs capsule structure as the backbone architecture. CNN-RNN~\cite{guera2018deepfake} is a sequence-based baseline method that uses CNN to generate a set of features from each frame and pass them to the LSTM. Finally, F3Net~\cite{qian2020thinking} is a frequency-based method that uses the Discrete Cosine Transform (DCT) to extract frequency domain information and analyze its statistical features for face forgery detection.

\paragraph{Evaluation Metrics}
\label{metric}
We apply the Accuracy score (ACC), Area Under the RoC Curve (AUC), and Equal error rate (EER) as our evaluation metrics, which are commonly used in the field of deepfake detection~\cite{zhao2021multi, nguyen2020eyebrow}.
\begin{itemize}
    \item AUC$ = \frac{\sum I\left ( P_{m}, P_{n} \right )}{M\ast N}$,
    $I\left ( P_{m}, P_{n} \right ) = \left\{\begin{matrix}
    1,   P_{m} > P_{n} \\
    0.5, P_{m} = P_{n} \\
    0,   P_{m} < P_{n} \\
    \end{matrix}\right.$,
    where $M$ and $N$ represent the number of positive and negative samples, respectively. In this paper, since we only consider the detection of fake videos, we use video-level AUC in the paper. The video-level AUC$ = \frac{\sum_{i=1}^n{\left(AUC_{i}\right)}}{n}$, where n represents the total number of selected frames in a video. Besides, $AUC_{i}$ means the AUC value for $i$-th frame.
    \item ACC$ = \frac{TP + TN}{TP + TN + FP + FN}$, where $TP$, $TN$, $FP$, and $FN$ are truly positive, true negative, false positive, and false negative, respectively.
    \item EER: the value when the false acceptance rate (FAR) is equal to the false rejection rate (FRR), where FAR$ = \frac{FP}{FP + TN}$, and FRR$ = \frac{FN}{TP + FN}$.
\end{itemize}
Note that we use video-level metrics for evaluation, which involve averaging the prediction results for each frame.

\subsection{Comparison with competing methods}
\label{comparisonwithbaseline}
To verify the effectiveness of our proposed framework, we conduct a comprehensive comparison with eight competing baseline methods in the same development environment and settings to ensure a fair comparison. 
These eight competing methods include HeadPose, FDFClassifer, Xception, MesoNet, Meso-Incep, CapsuleNet, CNN\_RNN, and F3Net. HeadPose and FDFClassifer are traditional detectors that utilize manual features for forgery detection, while Xception, MesoNet, Meso-Incep, and CapsuleNet are spatial-based methods that use CNNs to detect forgeries. CNN\_RNN is a temporal-based method that extracts temporal inconsistencies for detection, while F3Net is a frequency-based method that leverages information in the frequency domain for detection. 
% For each of these methods, we provide more detailed information in our supplementary material.
In contrast, our proposed approach is based on features from multiple modalities, including spatial, frequency, temporal, and facial landmarks. This comprehensive feature fusion enables our framework to outperform the existing methods in terms of forgery detection performance, as demonstrated in our experimental results.

The results presented in Tab.~\ref{baseline} show that our model achieves the overall best testing results among the eight baseline methods on both FF++ (c23) and FF++ (c40) in terms of three evaluation metrics, namely AUC, ACC, and EER. These results indicate that our proposed framework can effectively learn distinctive forgery features and achieve excellent performance on four manipulated datasets with both low and high compression levels.

Also, from this table, we can conclude that traditional detection methods based on manual features, such as HeadPose and FDFClassifier, have limited representation learning ability. Consequently, they are unable to accurately differentiate between genuine and fake videos. While detectors based solely on spatial features are capable of achieving good results for high-quality forgery videos, their performance declines when it comes to low-quality videos. Conversely, detectors that rely exclusively on frequency features are found to perform exceptionally well for highly compressed videos. 
% Our proposed method leverages a combination of spatial, frequency information, and temporal features, to achieve superior performance on both high and low-quality videos.

\subsection{Comparison with other competing methods}
\label{comparisonwithprevious}
In addition to comparing our method with competing baseline methods, we also compared it with other existing detectors on FF++ (c23) and FF++ (c40) benchmarks. The experimental results are directly cited from \cite{chen2021attentive}. As shown in Table \ref{previous}, our proposed model achieves the overall best experimental results compared to most previous methods on both FF++ (c23) and FF++ (c40), highlighting the superiority of our approach. Additionally, it is worth noting that our model only requires 32 frames in each video for training, compared to most existing methods which use 270 frames, indicating that our model also has the advantage of requiring fewer computational resources.

\begin{table}[htbp]
\centering
\caption{Comparison of our proposed method with other existing methods on the FF++ dataset with different compression levels. We used the AUC and ACC metrics for evaluation.
}
\setlength{\tabcolsep}{1.0mm}{
\begin{tabular}{c|c|c|c|c|c}
\toprule
\multirow{2}{*}{Methods} &
\multirow{2}{*}{Frames} &
\multicolumn{2}{c|}{FF++ (c23)} &
\multicolumn{2}{c}{FF++ (c40)} \\
\cline{3-6}
& & AUC & ACC(\%) & AUC & ACC(\%) \\
\midrule
Steg.Features~\cite{fridrich2012rich} & $270$ & $-$ & $70.9$ & $-$ & $56.0$ \\
LD-CNN~\cite{cozzolino2017recasting} & $270$ & $-$ & $78.5$ & $-$ & $58.7$ \\
Face X-ray~\cite{li2020face} & $270$ & $0.874$ & $-$ & $0.616$ & $-$ \\
Cozzolino \textit{et al.}~\cite{cozzolino2017recasting} & $270$ & $-$ & $78.5$ & $-$ & $58.7$ \\
Bayer \& Stamm~\cite{bayar2016deep} & $270$ & $-$ & $83.0$ & $-$ & $66.8$ \\
Rahmouni \textit{et al.}~\cite{rahmouni2017distinguishing} & $270$ & $-$ & $79.1$ & $-$ & $61.2$ \\
Xcep-ELA~\cite{gunawan2017development} & $270$ & $0.948$ & $93.9$ & $0.829$ & $79.6$ \\
Xcep-PAFilters~\cite{chen2017jpeg} & $270$ & $0.902$ & $87.2$ & $-$ & $-$ \\
Two-Branch~\cite{masi2020two} & $270$ & $0.991$ & $96.9$ & $0.911$ & $86.8$ \\
Efficient-B4~\cite{tan2019efficientnet} & $270$ & $0.992$ & $96.6$ & $0.882$ & $86.7$ \\
SPSL~\cite{liu2021spatial} & $100$ & $0.953$ & $91.5$ & $0.828$ & $81.6$ \\
MD-CSND~\cite{agarwal2021md} & $270$ & $0.993$ & $97.3$ & $0.890$ & $87.6$ \\
\midrule
\textbf{Ours} & $32$ & $\textbf{0.995}$ & $\textbf{97.8}$ & $\textbf{0.938}$ & $\textbf{91.7}$ \\
\end{tabular}
}
\label{previous}
\end{table}

\begin{table}[htbp]
\centering
\caption{
Comparison of the generalization ability of our proposed method with state-of-the-art methods in terms of AUC. These models are trained on FF++ and tested on CelebDF. Results of other methods are directly cited from~\cite{liu2021spatial,AayushiAgarwal2021MDCSDNetworkMC}.
}
\setlength{\tabcolsep}{5.0mm}{
\begin{tabular}{c|c|c}
\toprule
\multirow{1}{*}{Methods} &
\multirow{1}{*}{FF++} &
\multirow{1}{*}{CelebDF} \\
\midrule
Two-stream~\cite{zhou2017two} & $0.701$ & $0.538$ \\
Meso4~\cite{afchar2018mesonet}  & $0.847$ & $0.548$  \\
Meso4Inception4~\cite{afchar2018mesonet}  & $0.830$ & $0.548$  \\
HeadPose~\cite{yang2019exposing}  & $0.473$ & $0.546$  \\
FWA~\cite{li2018exposing}  & $0.801$ & $0.569$  \\
VA-MLP~\cite{matern2019exploiting}  & $0.664$ & $0.550$  \\
Xception-raw~\cite{rossler2019faceforensics++}  & $0.997$ & $0.482$  \\
Xception-c23~\cite{rossler2019faceforensics++}  & $0.997$ & $0.653$  \\
Xception-c40~\cite{rossler2019faceforensics++}  & $0.996$ & $0.655$  \\
Multi-task~\cite{lu2020multi} & $0.763$ & $0.543$\\
Capsule~\cite{nguyen2019capsule} & $0.966$ & $0.575$\\
DSP-FWA~\cite{li2018exposing} & $0.930$ & $0.646$\\
Face-XRay~\cite{li2020face} & $0.991$ & $0.742$\\
F3Net~\cite{qian2020thinking} & $0.981$ & $0.652$ \\
Two-Branch~\cite{masi2020two} & $0.932$ & $0.734$ \\
Efficient-B4~\cite{tan2019efficientnet} & $\textbf{0.997}$ & $0.643$ \\
SPSL~\cite{liu2021spatial} & $0.969$ & $0.769$ \\
MD-CSND~\cite{agarwal2021md} & $0.995$ & $0.688$ \\
STIL~\cite{gu2021spatiotemporal} & $0.971$ & $0.756$ \\
CFFs~\cite{yu2022improving} & $0.976$ & $0.742$ \\

\midrule
\bf{Ours}  & $0.995$ &  $\textbf{0.888}$  \\ 
\bottomrule
	\end{tabular}}
\label{crossdataset}
\end{table}

\begin{table}[tb!]
\centering
\caption{
Ablation study regarding the effectiveness of each module in our approach. The frequency module is denoted by "FM", the landmark graph learning module by "LGL", and the temporal graph learning module by "TGL". We train our model on the FF++ and test it on the CelebDF. The best results obtained are highlighted, with the AUC metric used for evaluation.
}
\scalebox{1.0}{
\begin{tabular}{c|c|c|c|c|c}
\toprule
ID & FM & LGL & TGL & FF++ & CelebDF \\
\midrule
1 & - & - & - & 0.987 & 0.786 \\
2 & \checkmark & - & - & 0.990 & 0.799 \\
3 & \checkmark & \checkmark & - & \textbf{0.997} & 0.823  \\
4 & \checkmark & - & \checkmark & 0.992 & 0.815 \\
5 & \checkmark & \checkmark & \checkmark & 0.995 & \textbf{0.888} \\
\bottomrule
\end{tabular}}
\label{tab:threemodules}
\vspace{-5pt}
\end{table}

\vspace{-10pt}

\subsection{Generalization Ability Evaluation}
The generalization problem is a significant challenge in deepfake detection, as it pertains to the decline in detection performance of a model faced with testing datasets that differ in distribution from the training data. To evaluate the generalization ability of our proposed model, we train it on the FF++ (c23) dataset and test it on CelebDF, which is commonly used in generalization experiments for deepfake detection. We compare the generalization performance with existing state-of-the-art detectors.

Tab.~\ref{crossdataset} presents the experimental results of current state-of-the-art methods and our model. We observe that most existing detectors can achieve high AUC results in within-dataset evaluations, where both the training and testing data are from FF++ (c23). However, their performance can drop dramatically in cross-dataset evaluations with CelebDF. In contrast, our proposed method still achieves satisfactory performance on CelebDF, suggesting that our model can learn more generalizable features than other methods. This is mainly because we utilize artifacts from multiple views to learn a comprehensive representation for forgery detection. This multimodal representation enables us to obtain more generalizable detection results.

\begin{figure}[htb]  %当前位置排版，放不下可以在本页顶部或者底部
      \centering  %居中
      \includegraphics[width=1.0\linewidth]{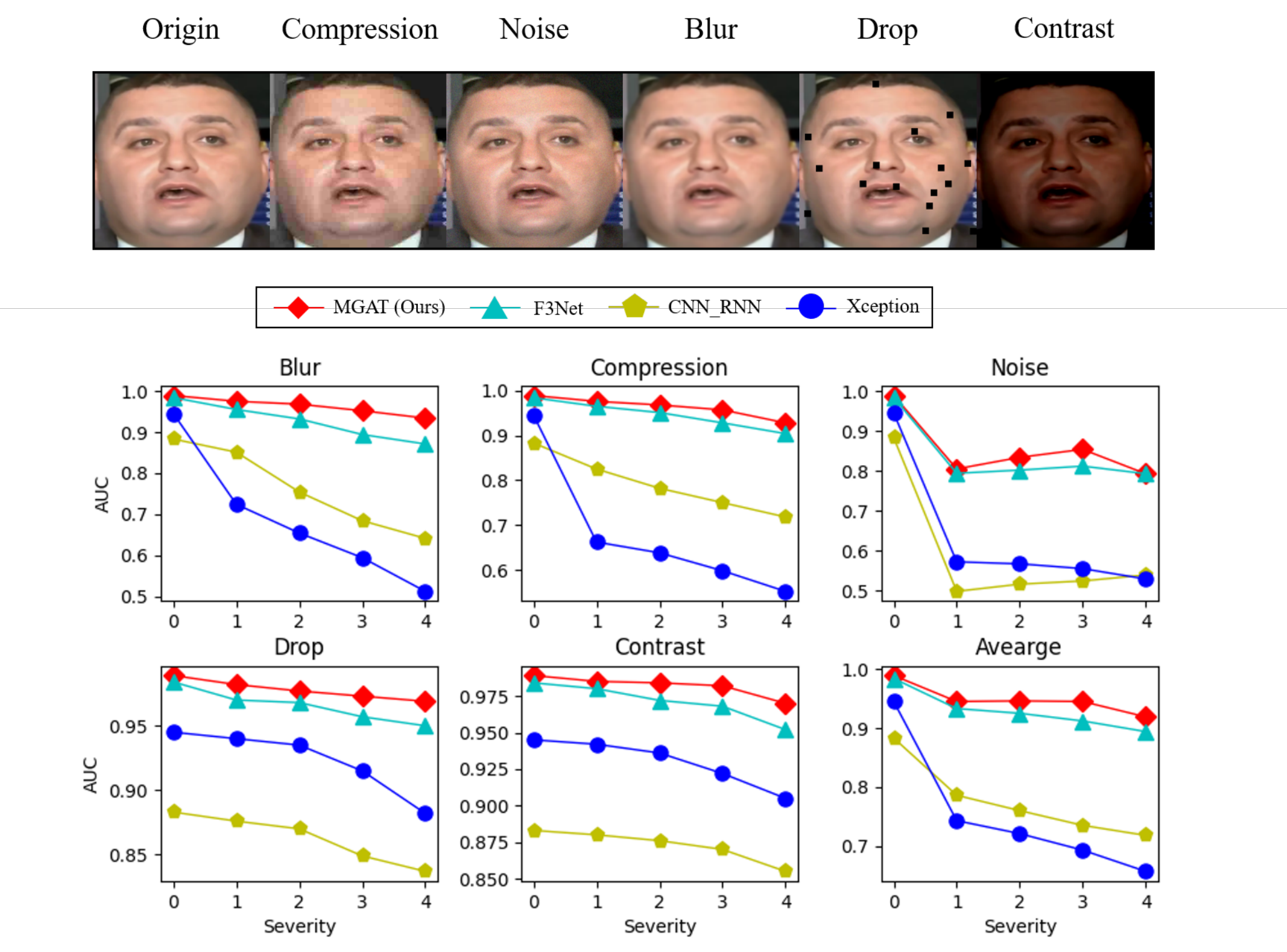} 
      \caption{
      Evaluation of the robustness of our method against unknown disturbances. The term "Average" in the figure denotes the average performance across all corruptions for each severity level.
      }
      \label{fig:data_aug} %指定给图片一个标签
\end{figure}

\subsection{Robustness Evaluation}
The robustness problem is a significant challenge in deepfake detection, as it involves the decline in the detection performance of a model faced with unseen disturbances. To evaluate the robustness of our proposed method, we compare it with three baselines: Xception, F3Net, and CNN\_RNN. The Xception model is based on spatial features, F3Net is based on frequency features, and CNN\_RNN is based on temporal features. The results shown in Fig.~\ref{fig:data_aug} demonstrate that our model can maintain detection performance, while other baselines experience a significant decline when faced with unknown disturbances. Additionally, we observe that detectors based solely on spatial features are most vulnerable, whereas frequency features are relatively more robust. By incorporating spatial, frequency, and temporal features, our model achieved the most robust results compared to other methods that only use one type of feature. This finding suggests that a multimodal approach can be an effective way to improve the robustness of deepfake detection models.

\begin{figure*}[htb]
      \centering
      \includegraphics[width=0.9\linewidth]{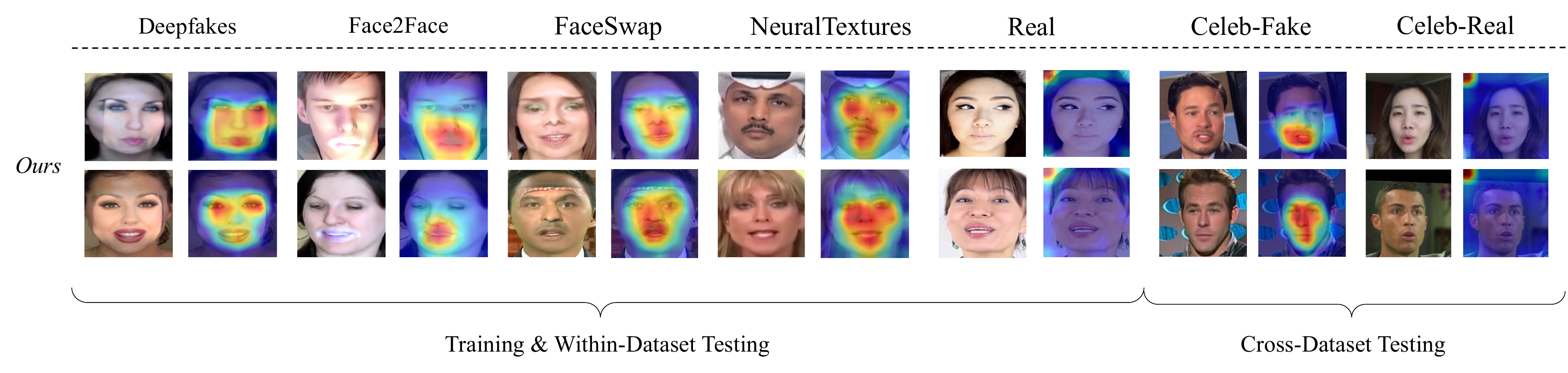} 
      \caption{
      Visualization of the performance of our proposed Multimodal Graph Learning (MGL) model trained on the FF++ dataset~\cite{rossler2019faceforensics++} and tested on the CelebDF dataset~\cite{li2020celeb}, which contains unseen data. We observe that our model accurately locates the forgery region and identifies the reasonable artifacts in the given forgery image, while correctly not responding to the real image. This demonstrates the effectiveness of our proposed detector and highlights the capability of our model in learning meaningful features in the images.
      }
      \label{fig:grad} %指定给图片一个标签
\end{figure*}

\subsection{Forgery Region Location}

% From Fig.~\ref{fig:visl}, we can find that the frequency-domain module tends to mine the texture features in the image, such as eyebrows, beard, hair, etc. On the contrary, the spatial-domain module tends to learn more semantic information about the face region, such as eyes, nose, etc. Based on this observation, we can show that the forgery features extracted in the frequency and spatial domain are mostly not the same. Therefore, combining these different forgery features can enhance the information content of training data, allowing the model to learn more robust knowledge.

To evaluate whether our model is able to locate forgery regions, we visualize the detection results using Grad-CAM~\cite{selvaraju2017grad}, as shown in Fig.~\ref{fig:grad}. Our model is trained on the four types of manipulation methods and tested on both within- and cross-datasets.
The visualization results demonstrate that our proposed model is able to capture meaningful artifacts of the face and detect most of the forgery regions. In contrast, the Xception baseline (shown in Fig.~\ref{fig:xcep_grad}) can be overfitted to the center region of a face and forgery-irrelevant regions, such as the background and the hair.
This visualization provides further evidence that our model is effective in detecting forgery and can locate the forgery regions in the manipulated videos. This also makes our model more interpretable, which is essential in practical applications.

\begin{table}[tb!]
\centering
\caption{
Comparison of the performance of different temporal modules, including the transformer, GRU, LSTM, RNN, and our proposed method. These models are trained on the FF++ dataset and tested on CelebDF, with the AUC metric used for evaluation. We also calculate the average scores obtained for each dataset.
}
\scalebox{1.0}{
\begin{tabular}{c|c|c|c|c}
\toprule
Method & FF++ & CelebDF & DFD & Avg \\
\midrule
RNN & 0.979 & 0.833 & 0.853 & 0.888 \\
GRU & 0.975 & 0.800 & 0.864 & 0.880  \\
LSTM & 0.990 & 0.853 & 0.866 & 0.904 \\
Transformer & 0.992 & 0.867 & 0.861 & 0.908 \\
Ours (Temporal GAT) & \textbf{0.995} & \textbf{0.888} & \textbf{0.904} & \textbf{0.929} \\
\bottomrule
\end{tabular}}
\label{tab:temporal}
\vspace{-5pt}
\end{table}

\begin{table}[tb!]
\centering
\caption{
Comparison of three feature fusion methods at the frame level. The first is the linear addition for feature fusion, which we denote as "Linear". The second is the fusion approach proposed by M2TR~\cite{wangm2tr}, which we denote as "M2TR". These methods are trained on the FF++ dataset and tested on CelebDF. The metric used in this table is AUC.
}

\scalebox{1.0}{
\begin{tabular}{c|c|c}
\toprule
Method & FF++ & CelebDF \\
\midrule
Linear & 0.969 & 0.834 \\
M2TR & 0.991 & 0.851  \\
Ours (SFF Module) & \textbf{0.995} & \textbf{0.888} \\
\bottomrule
\end{tabular}}
\label{tab:fusion}
\vspace{-5pt}
\end{table}

\begin{table}[tb!]
\centering
\caption{
Comparison of the performance of two landmark features on the FF++ dataset and tested them on CelebDF. The first method used only landmark coordinate sequences, while the second method obtained features using the GAT. We evaluated the performance of both methods using the AUC metric.
}
\scalebox{1.0}{
\begin{tabular}{c|c|c}
\toprule
Method & FF++ & CelebDF \\
\midrule
Only Lmk Seq & 0.988 & 0.814 \\
Ours (Lmk GAT) & \textbf{0.995} & \textbf{0.888} \\
\bottomrule
\end{tabular}}
\label{tab:landmark}
\vspace{-5pt}
\end{table}

\subsection{Ablation Study}
\label{ablationstudy}

To investigate the contribution of each component in our proposed model to the detection performance and generalization ability, we conduct a series of ablation studies. Our model comprises the following key modules: Frequency Module (FM), Landmark Graph Learning (LGL), and Temporal Graph Learning (TGL), as well as the Spatial-Frequency Fusion (SFF) method. We evaluate the effectiveness of each module in two experiments: one on a within-dataset and the other on a cross-dataset. Additionally, we compare our SFF with two other fusion methods, including linear addition and self-attention-based fusion proposed by M2TR~\cite{wangm2tr}. We also investigate the different representations of landmarks to verify the effectiveness of LGL.

First, we evaluate the effectiveness of the three modules (FM, LGL, and TGL) step by step on both within- and cross-dataset. For the baseline, we use the Xception model with pre-trained weights and commonly used augmentation methods. Results in Tab.~\ref{tab:threemodules} indicate that all three modules (FM, LGL, and TGL) contribute positively to the final prediction, and the addition of any of these components leads to better results on both within- and cross-dataset evaluations. Furthermore, the combination of all three modules results in significant improvement in generalization performance.

Second, we evaluate the effectiveness of our proposed temporal representation learning method, we compared it with four other sequential models: RNN, GRU, LSTM, and temporal Transformer. As shown in Tab.~\ref{tab:temporal}, our proposed method (temporal GAT) outperforms all other methods on all testing datasets, including FF++, CelebDF, and DFD. The transformer leverages a self-attention mechanism to learn both short- and long-term dependencies, making it perform better than the other three methods that can only capture short-term forgery features between adjacent frames. However, our proposed temporal GAT can learn both short- and long-term dependencies by computing the cosine similarity between two frames to encode the edge features. This approach allows us to explicitly learn the relationship between any given two frames and capture more complex temporal patterns in the videos. Therefore, the proposed temporal GAT is more effective than the other models in capturing the temporal dependencies in the video data, leading to more accurate and reliable deepfake detection.

Third, we evaluate the effectiveness of multimodal feature fusion at the frame level. The simplest way to fuse spatial and frequency features is to linearly add or concatenate them to obtain a unified representation at the frame level. Another fusion method proposed by M2TR~\cite{wangm2tr} utilizes the self-attention mechanism to treat spatial features as queries and frequency features as keys and values, which can obtain better results than linear addition. However, their assumption that spatial features are more important than frequency features does not always hold. To address this limitation, we propose a bi-directional self-attention mechanism and a gating mechanism to adaptively fuse spatial and frequency features. As shown in Tab.~\ref{tab:fusion}, our SFF module outperforms both the linear addition and M2TR in terms of detection performance. The main reason is that our SFF can dynamically learn the importance of spatial and frequency features, which can capture more comprehensive and discriminative information for forgery detection. In contrast, the linear addition lacks the ability to learn feature importance, while M2TR's assumption can limit its effectiveness on certain types of data.

Last, we evaluate the effectiveness of our proposed LGL module. Conventionally, facial landmarks are represented as a coordinate sequence that records only the 2D coordinate for each point of the face, thus resulting in a low-dimension representation. However, facial landmarks can be naturally treated as a graph. By using the GAT, we can obtain a more complex and high-dimension representation of landmarks. Additionally, we can explicitly learn the geometric dependencies between points of the face by using GAT. As shown in Tab.~\ref{tab:landmark}, after incorporating the LGL module, we achieve better results for both within- and cross-dataset evaluations. This demonstrates that our LGL module is necessary and can achieve better results than the method that only uses landmark coordinate sequences. By leveraging the graph structure of facial landmarks, our LGL module can effectively capture the relationship between different landmarks and improve the performance of forgery detection.

\vspace{10pt}

\section{LIMITATIONS}
Despite the promising results achieved by our proposed Multimodal Graph Learning (MGL) framework for deepfake detection, there are several limitations to consider. Firstly, the computational and memory requirements of our model are higher compared to other deepfake detection methods, which may make it unsuitable for real-time applications. Secondly, our framework is limited to detecting manipulated facial images and videos and may not be generalizable to other media types, such as audio. Lastly, the scalability of the backbone architecture used for training may affect the performance of our proposed model, which can limit its generalization and robustness. Due to limited computing resources, we are unable to explore this further through experiments.

\vspace{10pt}

\section{CONCLUSION}
This paper presents a Multimodal Graph Learning (MGL) framework designed to overcome the limitations of current deepfake detection methods by effectively extracting and utilizing distinguishing features for detection. Our framework leverages information from multiple modalities using two GNNs and several multimodal fusion modules. At the frame level, we utilize a bi-directional cross-modal transformer and an adaptive gating mechanism to fuse spatial and frequency features with the geometric-enhanced landmark features captured by a GNN. At the video level, we employ a graph attention network (GAT) to capture temporal inconsistencies between frames. Our proposed method captures both short-term and long-term patterns, as the GNN captures local information through feature aggregation and global information through a graph-level readout function. To the best of our knowledge, no existing work utilizes GNNs to model temporal information, suggesting that further exploration is necessary to determine the efficacy of GNN architecture in learning temporal forgery features.

Extensive experiments demonstrate that our proposed method significantly improves the generalization ability and robustness of deepfake detection by incorporating the Frequency Module (FM), Landmark Graph Learning (LGL), Temporal Graph Learning (TGL), and Spatial-Frequency Fusion (SFF) methods. Additionally, we evaluate the effectiveness of our proposed LGL and TGL modules and show that they contribute positively to the final prediction. Our proposed method achieves superior results on widely-used benchmarks, including the FF++, CelebDF, and DFD datasets.

Overall, our study suggests that combining information from multiple modalities can significantly improve the performance of deepfake detection and enhance its robustness and generalization abilities. Future research can explore more advanced techniques for incorporating additional modalities and further improving the performance of deepfake detection.

\clearpage

\bibliographystyle{IEEEtran}
% \bibliography{IEEEabrv,ref}
\bibliography{main}

\end{document}